\documentclass[conference]{IEEEtran}
\IEEEoverridecommandlockouts
\ifdefined\pdfobjcompresslevel\pdfobjcompresslevel=0\fi 
\usepackage[T1]{fontenc}
\usepackage{amsmath}
\usepackage{graphicx}
\usepackage{booktabs}
\usepackage{amsfonts}
\usepackage[table]{xcolor}
\usepackage{amssymb}
\usepackage{colortbl}

\definecolor{syntheticrow}{RGB}{232,242,252}
\definecolor{realrow}{RGB}{235,247,235}

\usepackage{tabularx}
\usepackage{multirow}

\newcolumntype{Y}{>{\raggedright\arraybackslash}X}

\usepackage{pifont}

\usepackage{caption}
\usepackage[colorlinks=true,urlcolor=blue,linkcolor=black,citecolor=black]{hyperref}

\begin{document}
\title{XCT-SAM: Sequential Parameter-Efficient Domain Adaptation of SAM for Industrial XCT Defect Segmentation}

\author{
\IEEEauthorblockN{
Md Mahedi Hasan\textsuperscript{1,*},
Md Mushfiqur Rahaman\textsuperscript{2,3,*},
Alan Pachkovskiy\textsuperscript{1},\\
Imtiaz Ahmed\textsuperscript{2},
Jeremy Dawson\textsuperscript{1},
Srinjoy Das\textsuperscript{1,2,3}
}
\IEEEauthorblockA{
\textsuperscript{1}Lane Department of Computer Science \& Electrical Engineering\\
\textsuperscript{2}Department of Industrial and Management Engineering\\
\textsuperscript{3}School of Mathematical and Data Sciences\\
West Virginia University, Morgantown, WV\\
\{mh00062, mr00131, ap00076\}@mix.wvu.edu,
\{imtiaz.ahmed, jeremy.dawson, srinjoy.das\}@mail.wvu.edu
}
\thanks{\textsuperscript{*}Equal contribution. Corresponding author: srinjoy.das@mail.wvu.edu}
\thanks{This paper has been accepted to the IAPR Workshop on Machine Vision for Industrial Inspection (MVI2) at the 28th International Conference on Pattern Recognition (ICPR 2026), Lyon, France.}
}

\maketitle

\begin{abstract}
Defect segmentation in additive manufacturing (AM) X-ray computed tomography (XCT) images remains challenging due to severe class imbalance and large distribution shifts across scan conditions. Although recent foundation models such as the Segment Anything Model (SAM) provide strong general-purpose segmentation priors, their natural-image pre-training transfers poorly to the AM XCT domain, where defects appear as subtle non-semantic microstructural anomalies. Moreover, adapting SAM to the AM domain is further limited by the large domain gap and scarcity of labeled real XCT data. We present \textbf{XCT-SAM}, a sequential parameter-efficient adaptation framework for AM XCT defect segmentation. Instead of adapting SAM directly from natural images to XCT data, we first fine-tune Conv-LoRA adapters on an alloy-microstructure dataset and subsequently transfer the adapted model to XCT images, progressively bridging the domain gap. Using Conv-LoRA adapters with rank $r=2$, the framework injects convolutional spatial inductive bias into SAM's backbone while training approximately 4.15M parameters and keeping over 99\% of the model frozen. We evaluate XCT-SAM on out-of-distribution CycleGAN-XCT benchmarks and real-world NIST XCT scans. Across both settings, XCT-SAM consistently outperforms zero-shot SAM and other domain-adapted SAM baselines, achieving the best overall IoU and Dice scores. These results demonstrate the effectiveness of intermediate domain adaptation with parameter-efficient adapters for industrial XCT defect segmentation. The source code is publicly available at \url{https://github.com/Mahedi-61/XCT-SAM.git}.
\end{abstract}


\begin{figure*}[t]
    \centering
	\includegraphics[width=0.9\textwidth]{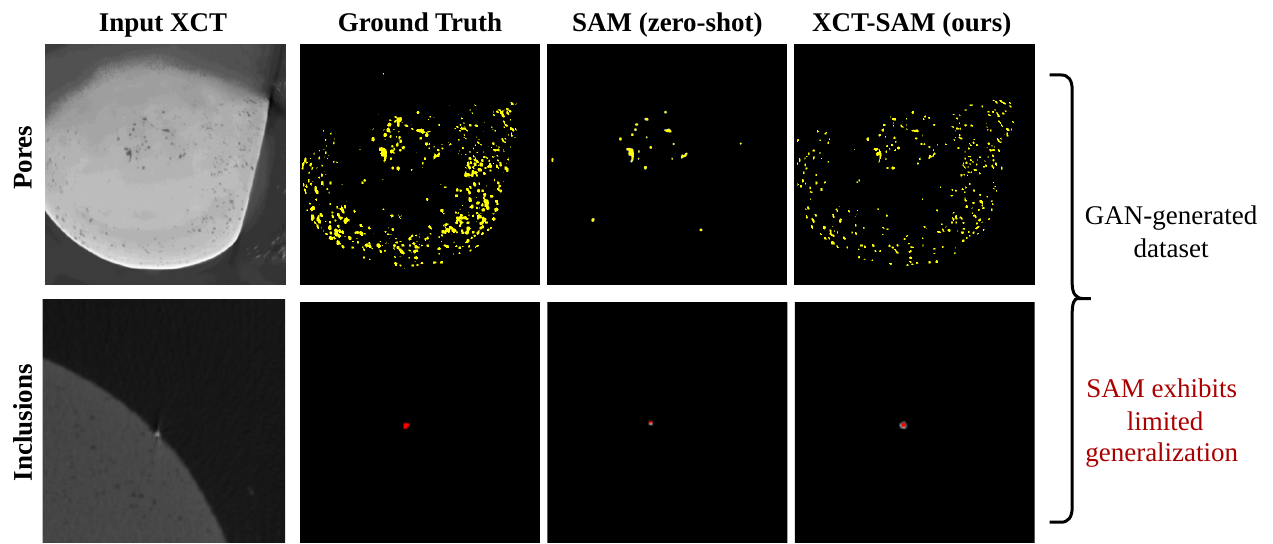}
	\caption{SAM exhibits limited generalization on AM XCT defect segmentation due to a significant domain gap. Our XCT-SAM addresses this through sequential domain adaptation combined with parameter-efficient fine‑tuning on limited XCT data, achieving enhanced performance and efficiency with only 0.647\% of trainable parameters.}
	\label{fig:figure_1}
\end{figure*}

\section{Introduction}
Additive manufacturing (AM) has emerged as a transformative manufacturing paradigm for producing geometrically complex, lightweight, and functionally optimized parts across aerospace, automotive, and biomedical applications \cite{ngo2018additive}. In contrast to conventional subtractive manufacturing, AM builds parts layer by layer directly from digital models, enabling high design flexibility and reduced material waste~\cite{wang2026recent}. These benefits make AM particularly suitable for high-value engineering parts with complex internal geometries ~\cite{wang2026recent,blakey2021metal}.

However, despite these advantages, the layer-wise nature of AM also introduces significant challenges in process stability and part quality. For instance, variations in laser power, scan speed, melt-pool dynamics, and environmental conditions can induce internal defects, including pores, cracks, lack-of-fusion regions, and inclusions~\cite{tusher2025comprehensive,haribaskar2024defects}. These defects are typically microscopic, sparsely distributed, and irregular in shape~\cite{poudel2022feature} and can severely affect mechanical strength and structural reliability~\cite{wang2026porosity,zhan2025assessment}. Since such defects are embedded within the part, they cannot be reliably identified through surface inspection alone~\cite{lapre2024rapid,bimrose2025detecting}. As a result, accurate internal defect segmentation is critical for AM quality assurance and certification. To address this challenge, X-ray computed tomography (XCT) has emerged as a widely adopted non-destructive testing (NDT) modality in AM~\cite{sun2025x,jones2025validation,baig2025non}. XCT enables high-resolution 3D visualization of internal structures and supports accurate segmentation of internal defects.

For automated defect segmentation in AM XCT images, traditional image-processing techniques such as thresholding and region growing often require extensive expert tuning and struggle to generalize across datasets with out-of-distribution materials~\cite{perghem2025ml,ledwaba2025development}. In contrast, deep learning approaches can learn complex fine-grained defect patterns from sufficiently diverse training data~\cite{unet++}. More recently, foundation models have emerged as a promising direction for AM defect segmentation~\cite{era2025unsupervised,tabassum_2024_adapting,gruber2024adapting,ma2025alloy}. In particular, the Segment Anything Model (SAM)~\cite{sam_2023,sam2_2024,medsam_2024,sam_med2d} and its variants have demonstrated strong capability in learning generalizable defect representations.

However, adapting SAM to the AM domain presents a core challenge due to a substantial distribution gap between the pre-training and target domains. SAM was originally trained on SA-1B, which contains color-rich images with high-level semantic structure \cite{sam_2023}. In contrast, AM XCT images consist of subtle low-contrast defect anomalies; inclusions are particularly sparse, occupying less than 0.4\% of the image area, while pore density varies considerably across datasets. In addition, directly fine-tuning SAM on AM XCT data is computationally expensive~\cite{tabassum_2024_adapting}, making it impractical for industrial real-world deployment.



The high cost of directly fine-tuning SAM arises from its image encoder (ViT-H), which alone comprises 636M parameters, and fine-tuning such a large model on limited training data risks overfitting. This motivates parameter-efficient fine-tuning (PEFT) methods, where lightweight trainable modules are injected into a frozen backbone while the majority of parameters remain fixed. A prominent PEFT approach is Low-Rank Adaptation (LoRA)~\cite{lora_2022}, which approximates weight updates through low-rank projections, substantially reducing the number of trainable parameters. Conv-LoRA~\cite{conv_lora_2024} extends this framework by augmenting each low-rank update with parallel convolutional experts, introducing spatial inductive bias that standard LoRA lacks, thereby enabling the model to capture fine-grained spatial details such as pore and inclusion boundaries. Conv-LoRA has recently been applied to AM XCT defect segmentation in~\cite{tabassum_2024_adapting}, demonstrating that synthetic supervision can be used to fine-tune SAM for this domain. However, adapting Conv-LoRA to the AM domain remains challenging due to the large distribution shift from SAM's natural-image pre-training, the need to represent fine-grained XCT defect patterns within a constrained low-rank space, irregular morphology, and severe class imbalance of AM defects.

To address these challenges, we propose XCT-SAM, a curriculum-inspired, sequentially adapted, parameter-efficient framework that progressively bridges the domain gap between natural images and AM XCT defect data for segmenting diverse defect types.
In the first stage, we fine-tune Conv-LoRA SAM on alloy-microstructure data before transitioning to XCT data. These microstructures share key properties with XCT images, including homogeneous textures and sparse, point-like defects. This stage shifts the LoRA parameters toward a feature space better aligned with AM characteristics. In the second stage, we initialize the Conv-LoRA-adapted SAM with the alloy-adapted weights. This warm start reduces the effective domain gap, enabling the limited XCT data to drive more stable optimization, faster convergence, and improved generalization.

\noindent The main contributions of this work are summarized as follows:
\begin{itemize}
    \item We introduce a \textbf{two-stage material-to-XCT adaptation strategy} that first adapts SAM on real alloy-microstructure and then transfers it to AM XCT defect data. This decomposes the large natural-image-to-XCT shift into two smaller adaptation steps, improving stability under scarce and highly imbalanced supervision.

    \item We instantiate this strategy as \textbf{XCT-SAM} by using Conv-LoRA adapters within a frozen SAM ViT-H backbone and optimizing their rank for AM XCT defect segmentation. The best configuration uses $r=2$ with eight convolutional experts, fine-tuning only 4.15M parameters, or 0.647\% of the full model.

    \item We validate XCT-SAM on both synthetic and real-world Out of Distribution (OoD) benchmarks, where it consistently outperforms zero-shot SAM and competitive fine-tuned baselines across defect segmentation tasks.
\end{itemize}


\section{Related Work}

\subsubsection{XCT Defect Segmentation in AM}
Prior works on AM XCT defect segmentation have shown that segmentation quality plays a critical role in downstream defect analysis. It was shown in~\cite{poudel2022feature} that XCT segmentation quality strongly affects defect detectability and size estimation in laser powder bed fusion parts. Their results indicate that U-Net-based segmentation can improve defect detection and sizing over commonly used thresholding methods such as Otsu in several settings. However, supervised segmentation models still depend on labeled XCT data and may require additional post-processing to recover irregular or fragmented defect patterns~\cite{baig2025non}. Moreover, their performance can degrade under variations in material, scan configuration, noise level, and defect morphology, motivating more data-efficient and domain-adaptive segmentation methods.

\subsubsection{SAM for AM XCT Segmentation}
Medical adaptations such as MedSAM~\cite{medsam_2024} and SAM-Med2D~\cite{sam_med2d} show that domain-specific fine-tuning can improve SAM on low-contrast images, and we include them as transfer baselines. However, AM XCT defect segmentation differs substantially from medical lesion segmentation: defects are sparse, low-contrast, non-semantic, and highly sensitive to material composition, scan parameters, reconstruction quality, and defect morphology. Recent studies have begun adapting SAM to AM XCT defect segmentation. Era et al.~\cite{era2025unsupervised} proposed an unsupervised prompt-generation framework for SAM-based porosity segmentation, reducing the need for manual annotation but remaining prompt-dependent under noisy backgrounds, heavy-tailed defect distributions, and complex morphologies. Tabassum and Ziabari~\cite{tabassum_2024_adapting} adapted SAM with Conv-LoRA on CycleGAN-generated XCT data and trained separate binary models for material, pore, and inclusion segmentation. Their study showed that synthetic XCT supervision improves over a 2.5D U-Net baseline~\cite{cheniour2024mesoscale}; however, generalization remains limited under out-of-distribution settings, and multiclass defect segmentation remains challenging under severe class imbalance.


\begin{figure*}[t]
	\includegraphics[width=\textwidth]{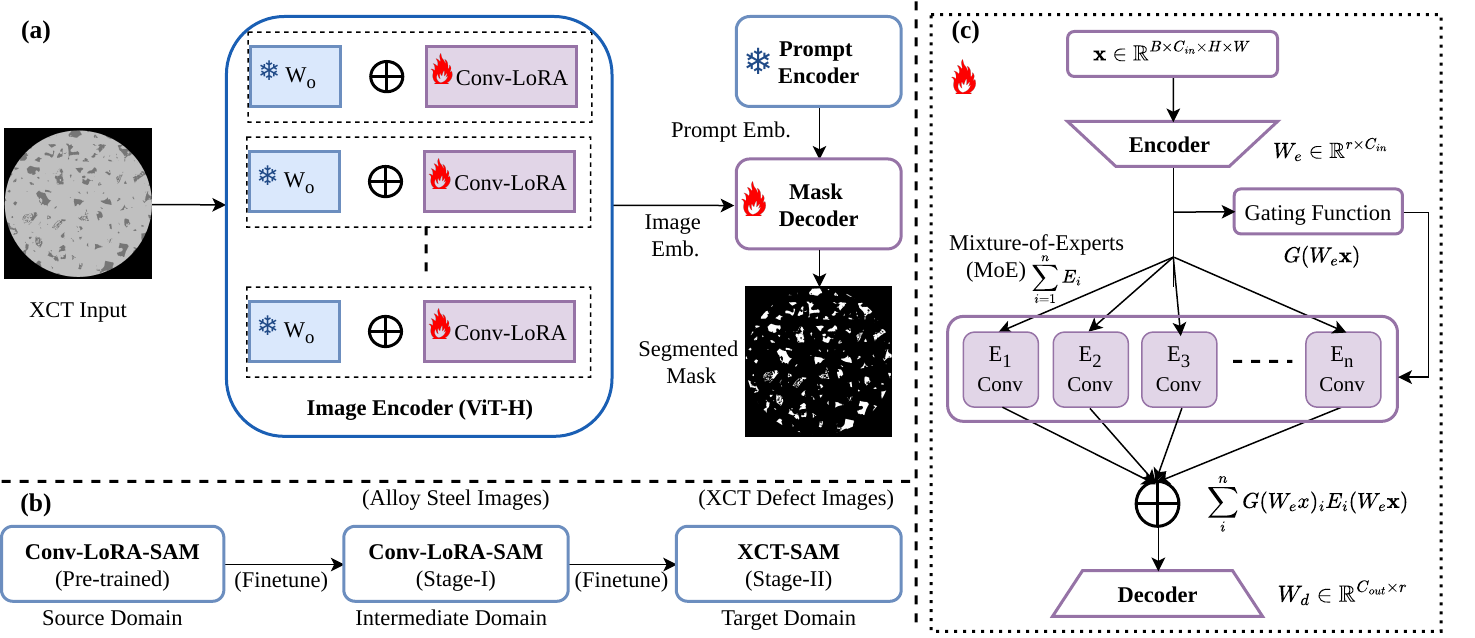}
    \caption{Overview of XCT-SAM. Conv-LoRA adapters are inserted into the frozen SAM ViT-H encoder, first adapted on alloy-microstructure images and then transferred to AM XCT defect segmentation.}
	\label{fig:overview_xct_sam}
\end{figure*}

\section{Methodology}

\subsection{Framework Overview}
Figure~\ref{fig:overview_xct_sam} illustrates the overall architecture of the proposed XCT-SAM framework for defect segmentation in XCT images. Our approach builds upon a SAM backbone based on the ViT-H image encoder ($\sim 636$M parameters), which provides a strong visual representation. We hypothesize that adapting SAM to the AM domain can be achieved through a low-dimensional domain-specific shift, where the model's attention is redirected from semantic objects to fine-grained defect patterns. The proposed XCT-SAM framework incorporates Conv-LoRA~\cite{conv_lora_2024}, a lightweight convolutional low-rank adaptation module, into each transformer block of the frozen ViT-H encoder. We keep the ViT-H backbone entirely frozen during training to preserve its learned pre-trained representations. Within the Conv-LoRA adapter, a set of convolutional expert gates is introduced between the encoder ($W_E$) and decoder projection ($W_D$) layers. Specifically, for each adapted layer, the embedding update is computed as:
\[
H = W_0 x + W_D \left( \sum_{i=1}^{N} g_i(x)\, E_i(W_E x) \right),
\]
where $W_0$ denotes the frozen backbone weight, $W_E$ and $W_D$ denote the trainable low-rank projection matrices, $E_i(\cdot)$ represents the $i^{th}$ convolutional expert $i=1,\ldots,N$, and $g_i(x)$ is the gating weight assigned by the gating network. Here, the rank $r \ll \min(C_{\mathrm{in}}, C_{\mathrm{out}})$ controls the trade-off between parameter efficiency and adaptation capacity. In this study, we evaluate three rank configurations, $r \in \{2,4,8\}$, to investigate this trade-off for AM defect segmentation.

Since fine-tuning is performed on a relatively small dataset ($\sim$1K images), higher ranks may increase the risk of overfitting or unstable minority-class adaptation. We therefore hypothesize that lower-rank adaptations are sufficient to capture the domain-specific shift while preserving generalization. Experimental results support this hypothesis, with $r=2$ achieving the best average segmentation performance across all evaluation benchmarks. Furthermore, this lower-rank configuration substantially reduces the number of trainable parameters while maintaining similar inference latency and memory consumption because computation is dominated by the frozen SAM backbone. To mitigate the severe class imbalance inherent in AM XCT images, we train two independent binary segmentation models, dedicated to pore and inclusion detection, respectively. We adopt this multi-model formulation because prior work reported that joint multi-class training under extreme imbalance leads to reduced generalization performance, particularly under out-of-distribution settings~\cite{tabassum_2024_adapting}.

\subsection{Objective Functions}

In this study, we evaluate multiple objective functions, treating loss design as a component of domain adaptation rather than a purely generic segmentation choice. This is important because AM XCT defects are not only statistically rare, but also physically meaningful structures whose missed detection, boundary distortion, or fragmentation can affect downstream defect analysis. Unlike prior Conv-LoRA-based SAM adaptation for AM XCT, which primarily used a structure loss combining BCE and IoU terms~\cite{tabassum_2024_adapting}, we explicitly study objectives that encode different defect-segmentation priorities under extreme class imbalance. We first consider binary cross-entropy (BCE) loss, which optimizes each pixel independently. However, in AM XCT images, BCE is heavily dominated by the vast number of background pixels, thereby diminishing the learning signal associated with rare and tiny defect regions.

We also optimize XCT-SAM with Dice loss~\cite{dice_2016}, which normalizes overlap by the combined size of predicted and ground-truth regions, improving robustness to class imbalance. Focal Tversky loss~\cite{tversky_2017,focal_tversky_2019} extends the Dice formulation by introducing asymmetric penalties for false positives and false negatives using the Tversky index, while its focal exponent emphasizes hard examples. This is especially relevant for AM inspection, where missing a small pore or inclusion may be more consequential than modest over-segmentation. We further evaluate the Lovász-Softmax loss~\cite{loavaz_softmax_18}, which directly optimizes a differentiable surrogate of the IoU objective through a convex extension of its set-based formulation. This objective is included to assess whether directly targeting region-overlap quality improves defect boundary recovery under sparse and irregular masks.

Finally, we adopt the Dice-Focal loss~\cite{dice_2016,focal_2017}, which combines region-overlap optimization with hard-example mining. The focal loss 
reduces the contribution of easy background pixels and focuses training on sparsely distributed defect regions. Together, the Dice and focal terms provide stable gradients from defect pixels, even when they occupy only a tiny fraction of the image. Thus, our objective-function study connects the statistical challenge of class imbalance with the physical requirements of XCT defect segmentation: detecting small rare defects, preserving irregular morphology, and improving robustness across domain-shifted XCT data.

\section{Experiments}

\subsection{Datasets}
\label{sec:dataset}
In our experiments, we use both CycleGAN-generated synthetic XCT data~\cite{tabassum_2024_adapting} and real NIST XCT data~\cite{kim_2017_pore}. The CycleGAN-XCT dataset is generated by translating physics-based simulated XCT volumes into realistic XCT images while preserving pore and inclusion labels. Since only the translated synthetic images and corresponding annotations are publicly available, we use this dataset as the primary source for fine-tuning SAM to XCT defect segmentation. Moreover, the synthetic dataset provides a controllable alternative to real XCT data, which remains scarce and often fails to capture the full diversity of defect characteristics~\cite{tabassum_2024_adapting}.

To evaluate cross-distribution generalization on real XCT images, we use the NIST XCT dataset from the CoCr laser powder bed fusion study as described in~\cite{kim_2017_pore}. This dataset contains XCT scans of AM specimens acquired under varying process parameters, leading to diverse pore distributions and morphologies.
Since manually annotated ground-truth defect masks are not publicly available, we generate threshold-derived binary reference masks following the image-processing pipeline of ~\cite{kim_2017_pore}. As the NIST dataset contains only pore defects, inclusion segmentation evaluation is not applicable and is therefore omitted from the NIST benchmark results.

\begin{figure}[t]
    \centering
    \includegraphics[width=\columnwidth]{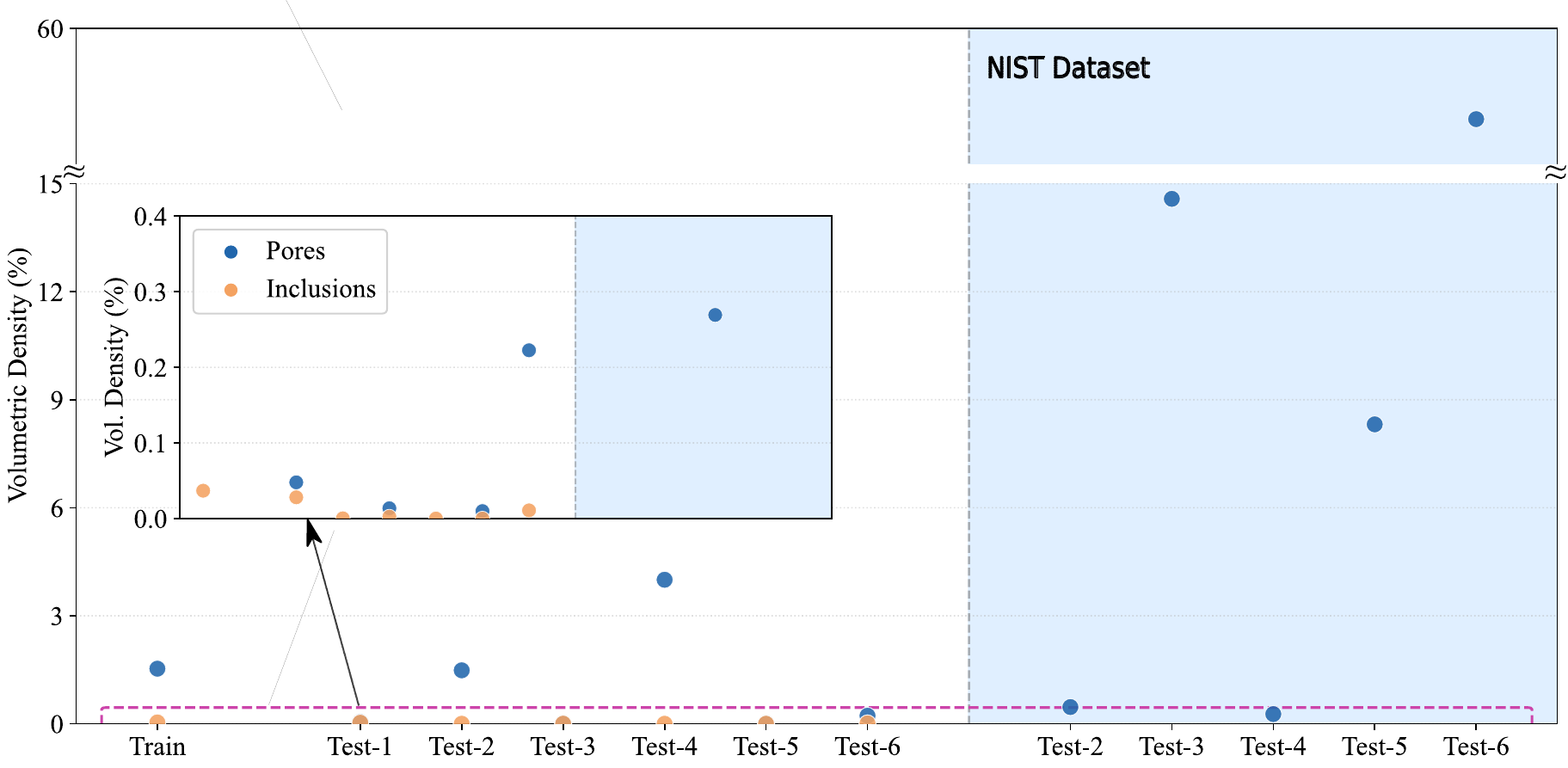}
    \caption{Comparison of volumetric defect density of the XCT data. In the training data, pore density reaches up to 3\%, whereas inclusion density remains below 1\%. Across all test sets, inclusion density is consistently lower than 0.3\% while pore density reaches up to 5\% in the GAN-generated data and up to 55\% in the NIST data.}
    \label{fig:density_plot}
\end{figure}

\begin{figure*}[t]
    \centering
    \includegraphics[width=0.85\textwidth]{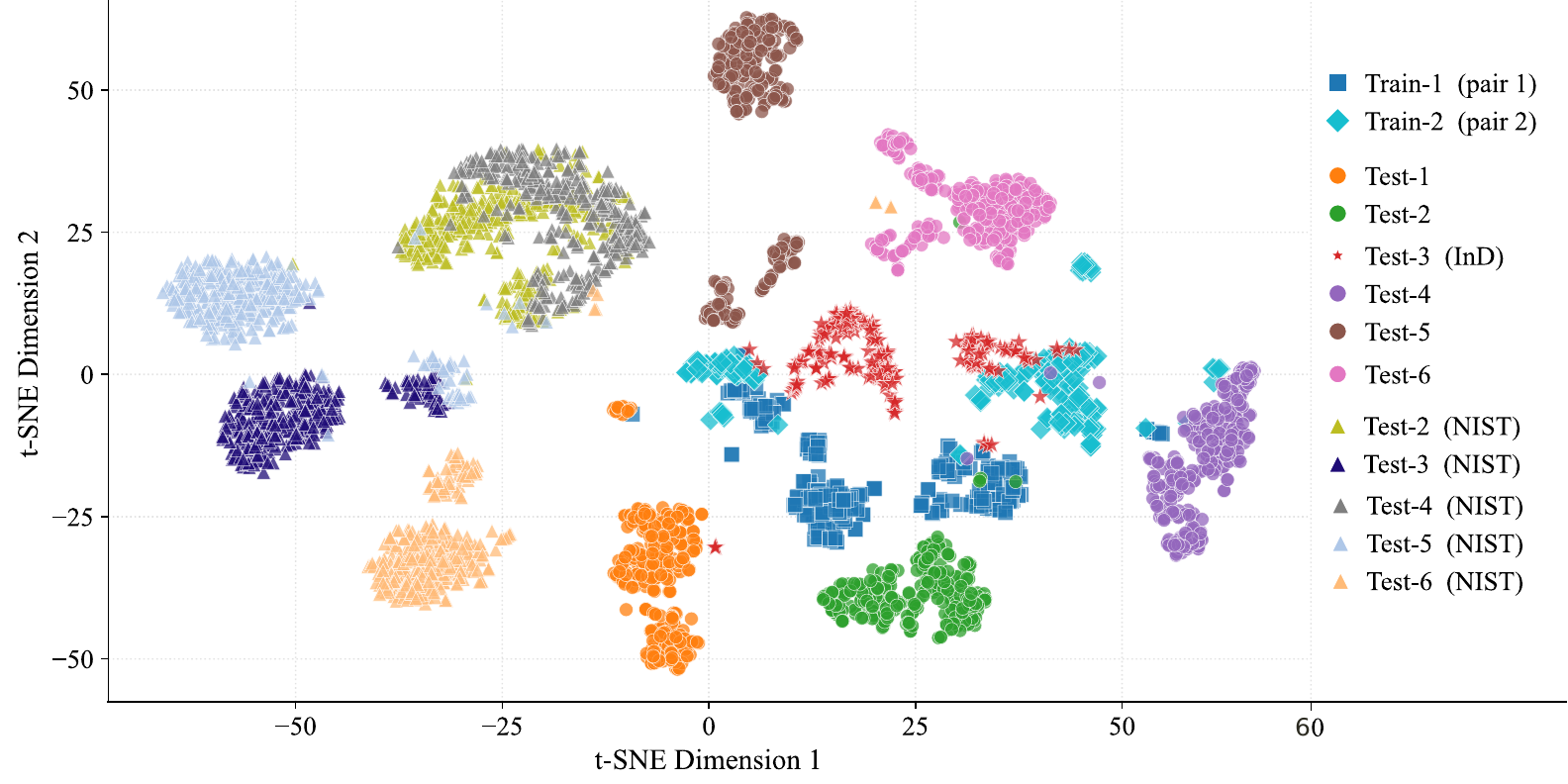}
    \caption{t-SNE visualisation of ViT-B/16 feature distributions across datasets.}
    \label{fig:tsne_plot}
\end{figure*}

Figure~\ref{fig:density_plot} illustrates variation in defect-density statistics across the training and evaluation sets. In the GAN-generated synthetic training data, pore density reaches up to approximately 3\%, whereas inclusion density remains below 1\%. Across the test sets, inclusions become even sparser, below 0.3\% across all sets, while pore density varies considerably and reaches up to approximately 5\%. In contrast, pore density in the NIST dataset reaches up to 55\%. This wide variation in defect density introduces severe class imbalance and makes AM XCT segmentation fundamentally different from natural object segmentation, where target objects typically occupy larger and semantically coherent regions.

We further analyze the feature-space distributions of the GAN-generated and NIST XCT datasets using the t-SNE visualization in Fig.~\ref{fig:tsne_plot}. As shown in Fig.~\ref{fig:tsne_plot}, \textit{Train-1}, \textit{Train-2}, and \textit{Test-3} cluster closely due to similar material properties and scan settings~\cite{tabassum_2024_adapting}, motivating the use of \textit{Test-3} as the validation set. In contrast, the remaining GAN-generated test sets are more dispersed, indicating synthetic out-of-distribution shifts caused by variations in scan settings and noise characteristics~\cite{tabassum_2024_adapting}. The NIST test sets form separate clusters far from the GAN-XCT data, highlighting a stronger synthetic-to-real domain gap caused by differences in acquisition protocol, and label-generation procedure. Accordingly, we use the remaining GAN-generated sets, \textit{Test-1}, \textit{Test-2}, and \textit{Test-4--Test-6}, and all NIST sets as synthetic and real-world OoD benchmarks, respectively.

For Stage-I adaptation, we use an \textbf{alloy-microstructure dataset}~\cite{alloy_2022} as an intermediate domain between natural images and AM defect data. Although not AM-specific, this dataset exposes the model to real metallic texture and microstructural patterns before final adaptation to AM XCT defects. 

\subsection{Baseline Models for Comparison}
We compare XCT-SAM against five baselines spanning supervised convolutional segmentation, zero-shot SAM, medical-domain SAM adaptation, and parameter-efficient SAM fine-tuning: UNet++, SAM, MedSAM, SAM-Med2D, and Conv-LoRA-SAM. UNet++~\cite{unet++} serves as a supervised CNN baseline with nested dense skip connections between encoder and decoder layers. We also evaluate the original SAM~\cite{sam_2023} in a zero-shot setting without any XCT-specific adaptation. 
Conv-LoRA-SAM~\cite{conv_lora_2024} integrates convolutional low-rank adaptation modules into SAM for efficient domain adaptation. Following the direct synthetic-XCT adaptation setting of~\cite{tabassum_2024_adapting}, we fine-tune this baseline on GAN-generated XCT images without the intermediate alloy-microstructure~\cite{alloy_2022} adaptation stage used in XCT-SAM. For fair comparison, all trainable baselines are optimized using the same XCT training data, preprocessing pipeline, and post-processing procedures where applicable. SAM is evaluated in a zero-shot setting, while all prompt-based methods use an identical prompt protocol across experiments.

\begin{figure*}[t]
	\includegraphics[width=\textwidth]{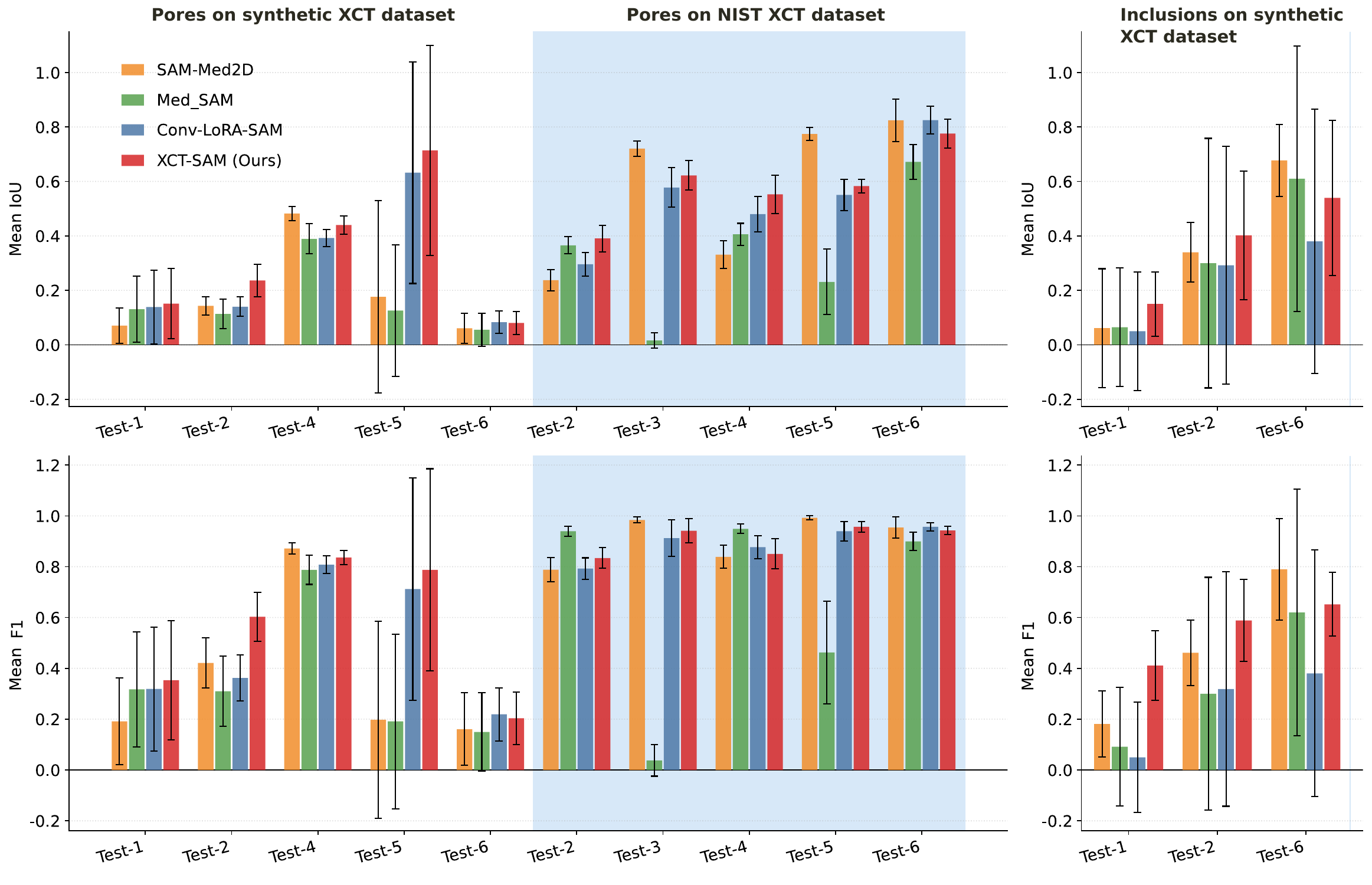}
	\caption{Comparison of segmentation performance across fine-tuned foundation models: SAM-Med2D, MedSAM, Conv-LoRA-SAM, and our XCT-SAM on GAN-generated XCT and NIST datasets. The top row shows mean IoU, while the bottom row shows the mean F1. The shaded blue region represents the NIST evaluation sets, and the error bars indicate $\pm 1$ std.}
	\label{fig:model_comparison}
\end{figure*}

\subsection{Evaluation Metrics}
We evaluate XCT-SAM using pixel-level metrics to measure segmentation accuracy. We first report Intersection over Union (\textbf{IoU}), which measures overlap quality while penalizing both over- and under-segmentation. As IoU is sensitive to small boundary shifts, we also compute Precision and Recall to quantify false positives and false negatives. We further report \textbf{Tolerance-F1}, which considers predictions within a 5-pixel radius of the ground truth as correct. Finally, to better handle severe class imbalance in defect segmentation, we compute the \textbf{Dice} coefficient, which provides a more robust measure of partial overlap.

\subsection{Implementation Details of XCT-SAM}
We fine-tune XCT-SAM in two stages using AdamW with polynomial learning-rate decay. The ViT-H SAM backbone is adapted via Conv-LoRA with rank $r=2$ and eight convolutional experts, and trained using the \textbf{Dice--Focal} loss. In Stage 1, we perform intermediate domain adaptation on the alloy dataset~\cite{alloy_2022} using a learning rate of $2\times10^{-4}$, a batch size of 8, and trained for up to 15 epochs. In Stage 2, we further adapt the model to the target AM XCT domain using a reduced learning rate of $8\times10^{-5}$ for an additional 20 epochs while keeping the same batch size.


\subsection{Quantitative Evaluation}
We evaluate XCT-SAM against five competitive baseline models across both GAN-generated and real NIST evaluation datasets, as illustrated in Fig.~\ref{fig:model_comparison} and summarized in Table~\ref{table:eval_table_1}. Our XCT-SAM achieves the best IoU and Dice performance across the three evaluation settings: pore segmentation on GAN-generated XCT, inclusion segmentation on GAN-generated XCT, and pore segmentation on real NIST XCT. This demonstrates improved generalization under both synthetic and real XCT imaging conditions.

For pore segmentation on synthetic images, XCT-SAM achieves the highest scores across all evaluation metrics, outperforming the next best method, Conv-LoRA-SAM, by a substantial margin of $+0.0472$ IoU, $+0.0549$ Dice, and $+0.0726$ F1. In contrast, the pre-trained SAM model in its zero-shot setting performs poorly on GAN-generated pores. This result highlights the significant domain gap between natural and XCT images. UNet++~\cite{unet++}, fine-tuned MedSAM~\cite{medsam_2024}, and fine-tuned SAM-Med2D~\cite{sam_med2d} provide only marginal improvements over zero-shot SAM. 
These results indicate that these direct fine-tuned baselines remain limited in capturing the low-contrast characteristics of AM defect pores.

\begin{table*}[t]
	\centering
	\footnotesize
	\caption{Quantitative comparison of mean segmentation performance computed across all test sets for pore and inclusion detection on GAN-generated XCT and NIST benchmark datasets.}
	\label{table:eval_table_1}
	\setlength{\tabcolsep}{1.3pt}
	\begin{tabular}{l@{\hspace{1mm}} ccc | ccc | ccc} \hline
		Methods   &\multicolumn{3}{c}{Pores(GAN)} &\multicolumn{3}{c}{Inclusions (GAN)} &\multicolumn{3}{c}{Pores (NIST)} \\\cmidrule(lr){2-4} \cmidrule(lr){5-7}  \cmidrule(lr){8-10}
		&IoU &Dice &F1     &IoU &Dice &F1    &IoU &Dice &F1 \\\hline
	    \rowcolor{gray!12}SAM~\cite{sam_2023} &0.1427 &0.1567  &0.1713  &0.2951  &0.3017  &0.3221  &0.4942 &0.6549   &0.8887 \rule{0pt}{2.5ex}\\
        UNet++~\cite{unet++} &0.1403  &0.2138  &0.3395  &0.1017  &0.1600   &0.3472  &0.1340  &0.2153  &0.3512\rule{0pt}{2.5ex}\\
		MedSAM~\cite{medsam_2024} &0.1630   &0.2464   &0.3511  &0.3248  &0.3287 &0.3371   &0.3381   &0.4607   &0.6577\rule{0pt}{2.5ex}\\
        SAM-Med2D~\cite{sam_med2d} &0.1868   &0.2648   &0.3687  &0.3597   &0.3972 &0.4773  &0.5776  &0.6978 &0.9116 \rule{0pt}{2.5ex}\\
        Conv-LoRA-SAM~\cite{conv_lora_2024} &0.2773 &0.3701 &0.4841  &0.2407 &0.2443 &0.2494 &0.5460 &0.6886 &0.8959\rule{0pt}{2.5ex}\\
		\rowcolor{blue!12} XCT-SAM (ours) &0.3245  &0.4250  &0.5567 &0.3639  &0.4055  &0.5513 &0.5849  &0.7288   &0.9051\rule{0pt}{2.5ex}\\\hline
	\end{tabular} \vspace{-2mm}
\end{table*}

XCT-SAM also achieves the best overall performance for inclusion segmentation on the GAN-generated XCT evaluation sets. A notable observation is the substantial performance degradation of Conv-LoRA-SAM on inclusions (IoU: $0.2407$, F1: $0.2494$), despite its relatively strong pore segmentation results. This suggests that the original Conv-LoRA formulation overfits to the dominant class and fails to generalize effectively to rare defects. Similarly, UNet++ exhibits severe performance collapse on inclusions, likely due to the heavy class imbalance in the training data. Although zero-shot SAM, fine-tuned MedSAM, and fine-tuned SAM-Med2D achieve moderate inclusion IoU scores, their F1 scores remain comparatively lower than XCT-SAM, indicating limited segmentation consistency. Finally, on the real NIST benchmark, XCT-SAM achieves the best IoU and Dice scores among all baselines. Interestingly, zero-shot SAM attains a relatively competitive IoU score, likely due to the more distinguishable pore morphology in real XCT scans. In contrast, UNet++ degrades significantly under this distribution shift, highlighting its limited cross-domain generalization.


\begin{table*}[t]
	\centering
	\footnotesize
	\caption{Ablation study on the effect of Conv-LoRA rank selection for parameter-efficient adaptation on the GAN-generated synthetic dataset. We evaluate different low-rank configurations in terms of trainable parameters, inference efficiency, and segmentation performance. Results show that the r=2 configuration achieves the best trade-off between efficiency and performance.}
	\label{table:abs_obj_rank}
	\setlength{\tabcolsep}{2pt}
	\begin{tabular}{l@{\hspace{1mm}} cccc | ccc | ccc} \hline
		Rank &Train. & Percent &Infer.  &Infer.  &\multicolumn{3}{c}{Pores (GAN)} &\multicolumn{3}{c}{Inclusions (GAN)} \\
		&Params &Train. &(s/img) &Memory &IoU &Dice &F1     &IoU &Dice &F1\\\hline
	    r=8  &5.27 M &0.821\% &3.571 &2450MB  &0.3171  &0.4068   &0.5342  &0.1991   &0.2436   &0.4216 \rule{0pt}{2.5ex}\\
        r=4  &4.50 M &0.702\% &3.690 &2447MB  &0.3101  &0.4010   &0.5305  &0.1848   &0.2273   &0.4022\rule{0pt}{2.5ex}\\
\rowcolor{blue!12} r=2   &4.15M &0.647\% &3.656 &2445MB &0.3245 &0.4250 &0.5567 &0.3639    &0.4055  &0.5513\rule{0pt}{2.5ex}\\\hline
	\end{tabular}
\end{table*}

\begin{table*}[t]
	\centering
	\footnotesize
	\caption{Ablation study comparing various state-of-the-art objective functions for pore and inclusion segmentation. Reported values represent mean segmentation performance averaged across all test sets.}
	\label{table:abs_obj_function}
	\setlength{\tabcolsep}{2pt}
	\begin{tabular}{l@{\hspace{1mm}} ccc | ccc | ccc} \hline
		Objective   &\multicolumn{3}{c}{Pores (GAN)} &\multicolumn{3}{c}{Inclusions (GAN)} &\multicolumn{3}{c}{Pores (NIST)} \\\cmidrule(lr){2-4} \cmidrule(lr){5-7}  \cmidrule(lr){8-10}
		Function&IoU &Dice &F1     &IoU &Dice &F1    &IoU &Dice &F1 \\\hline
	    BCE   &0.3189  &0.4111 &0.5516 &0.2193   &0.2524  &0.3555  &0.5026   &0.6464   &0.8365\rule{0pt}{2.5ex}\\
        Dice  &0.3329  &0.4285 &0.5645 &0.3000   &0.3200  &0.3900  &0.3990  &0.5253  &0.7524 \rule{0pt}{2.5ex}\\
		Lovász-Softmax~\cite{loavaz_softmax_18} &0.3219   &0.4158   &0.5604  &0.2291   &0.2728   &0.4134   &0.3868   &0.5150   &0.7454 \rule{0pt}{2.5ex}\\
		Focal Tversky~\cite{tversky_2017,focal_tversky_2019} &0.3390   &0.4334  &0.5686 &0.3452 &0.3671 &0.4569 &0.5400  &0.6767 &0.9093 \rule{0pt}{2.5ex}\\
		\rowcolor{blue!12} Dice-Focal (ours) &0.3245  &0.4250  &0.5567 &0.3639  &0.4055  &0.5513 &0.5849  &0.7288   &0.9051 \rule{0pt}{2.5ex}\\\hline
	\end{tabular}
\end{table*}


\subsection{Ablation Study}

\subsubsection{Effect of Conv-LoRA Rank}
Table~\ref{table:abs_obj_rank} presents an ablation study examining the effect of Conv-LoRA rank on parameter efficiency and segmentation accuracy. In this study, we evaluate three rank configurations: $r\in \{8, 4, 2\}$.  For each configuration, we also measure trainable parameter count, inference latency, and memory consumption. From Table~\ref{table:abs_obj_rank}, we observe that as the rank decreases, the number of trainable parameters drops from 5.27M ($0.821\%$) at $r=8$ to 4.15M ($0.647\%$) at $r=2$. In contrast, inference time and memory consumption remain nearly unchanged, with less than $3\%$ variation across ranks. This is expected since the LoRA adapters account for less than $1\%$ of the total parameters, while computation is dominated by the frozen SAM backbone.
Overall, the experimental results show that $r=2$ achieves the best trade-off between model efficiency and segmentation performance. In contrast, higher ranks lead to a noticeable drop in inclusion-segmentation accuracy, suggesting that over-parameterized LoRA adaptation may reduce generalization.

\begin{figure*}[t]
    \centering
	\includegraphics[width=0.85\textwidth]{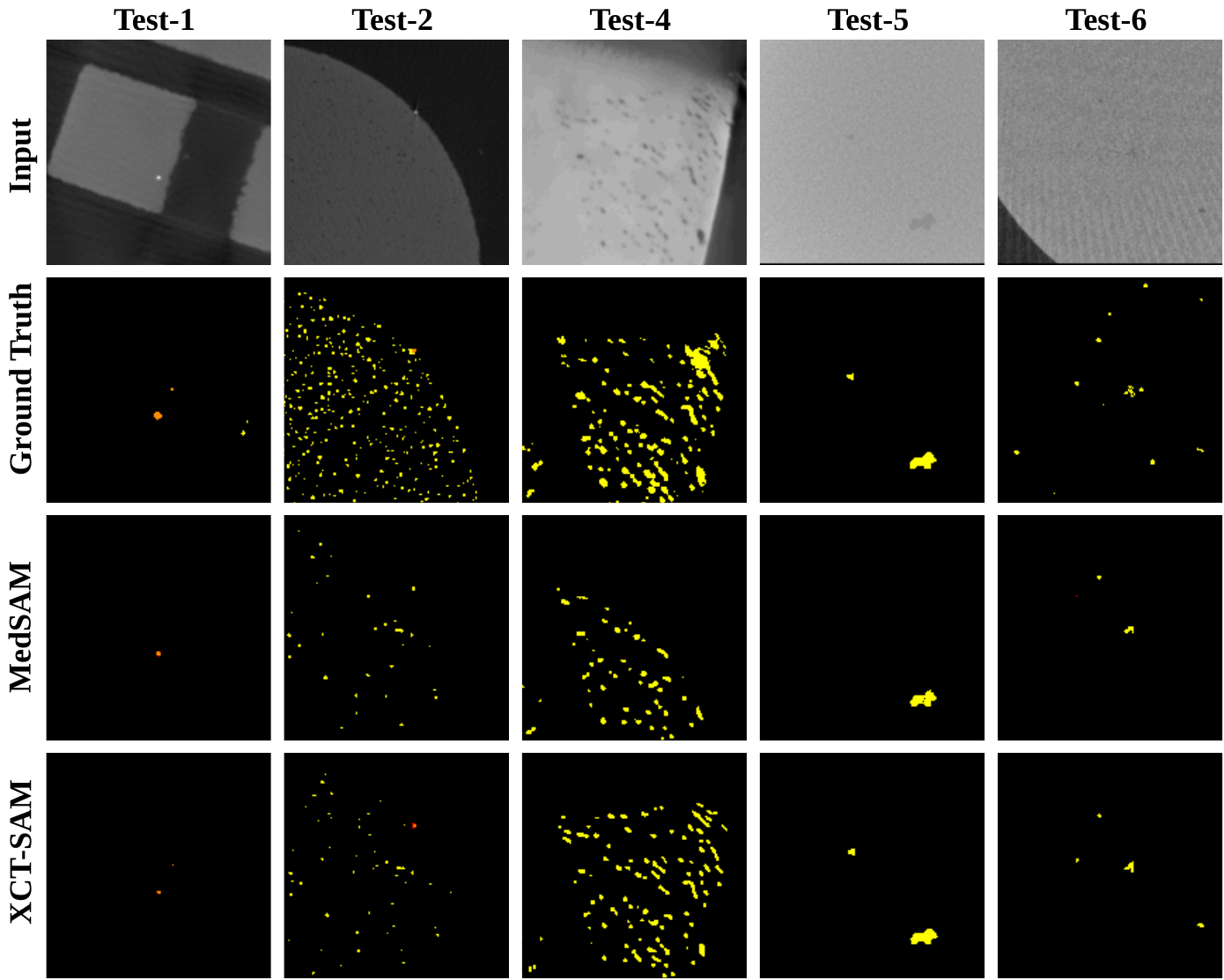}
    \caption{Qualitative comparison on GAN-generated XCT test sets. Yellow and red denote pores and inclusions, respectively. XCT-SAM better matches the reference masks than MedSAM across the shown OoD examples.}
	\label{fig:result_plot}
\end{figure*}

\begin{figure*}[t]
    \centering
	\includegraphics[width=0.85\textwidth]{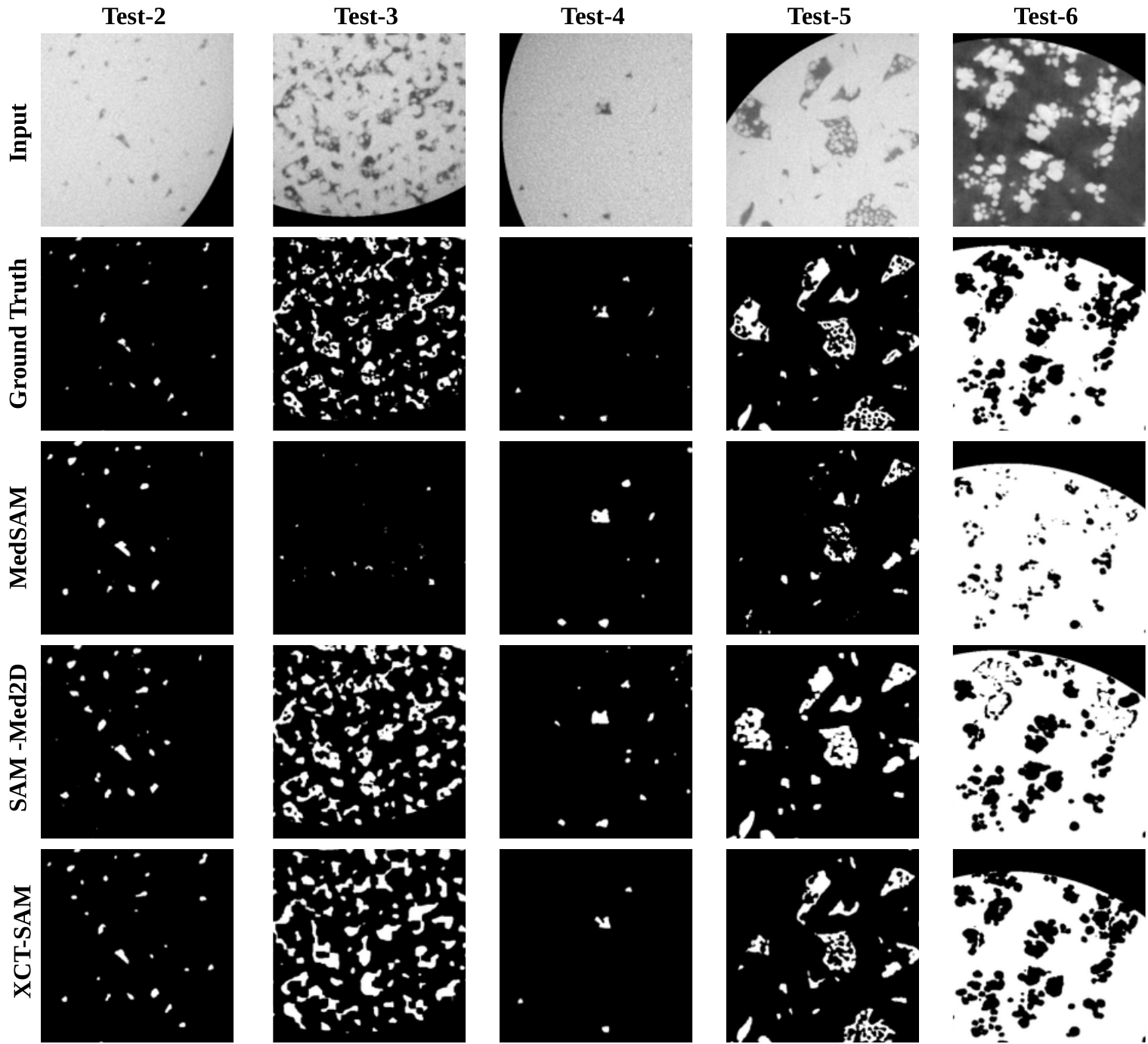}
    \caption{Qualitative comparison on NIST XCT test sets. The row marked Ground Truth denotes threshold-derived reference masks. XCT-SAM more closely matches the reference masks across varying defect densities.}
	\label{fig:nist_vis_comp}
\end{figure*}

\subsubsection{Effect of Objective Function}
Table~\ref{table:abs_obj_function} reports the effect of different loss functions for defect segmentation. The results yield several key observations. First, binary cross-entropy (BCE) achieves competitive performance on the majority class (pore). However, its performance degrades significantly on the minority class (inclusion), achieving only $0.2193$ IoU and $0.3555$ F1 score on the GAN-generated dataset. This behavior highlights the well-known sensitivity of BCE to severe class imbalance. Second, Dice loss improves inclusion segmentation compared to BCE by directly optimizing region overlap. However, this gain comes at the cost of degraded pore performance on the NIST dataset. Third, although Lovász-Softmax is designed to optimize IoU, it underperforms across all evaluation benchmarks in our experiments. In contrast, Focal Tversky loss achieves substantially enhanced inclusion segmentation by explicitly emphasizing minority classes, improving the inclusion IoU to $0.3452$. Nevertheless, its Dice and F1 scores for inclusions remain lower than those obtained with Dice-Focal objective.

Among all loss functions, Dice-Focal provides the most balanced performance across evaluation benchmarks and metrics. On the GAN-generated dataset, it achieves the best inclusion segmentation results, outperforming Focal Tversky by $+0.0187$ IoU and $+0.0944$ F1 score. On the NIST dataset, it also achieves the highest pore-segmentation performance with $0.5849$ IoU and $0.7288$ Dice. Overall, these results indicate that Dice-Focal provides robust generalization across out-of-distribution evaluation sets.

\subsection{Visualization}
We visualize sample segmentation results of XCT-SAM and representative SAM-based baselines on synthetic and real XCT scans in Fig.~\ref{fig:result_plot} and Fig.~\ref{fig:nist_vis_comp}, respectively. As shown in Fig.~\ref{fig:result_plot}, MedSAM often under-segments defects, whereas XCT-SAM produces predictions that remain well-aligned with the reference masks. For example, in \textit{Test-2}, MedSAM captures only a subset of dense pore regions and fails to recover many smaller pores. In contrast, XCT-SAM reconstructs a substantially larger portion of the pore distribution with enhanced spatial coverage and fewer false negatives. Similarly, in \textit{Test-6}, XCT-SAM better preserves fine defect structures than competing methods. Figure~\ref{fig:nist_vis_comp} further demonstrates the robustness of XCT-SAM under cross-distribution evaluation on real XCT scans. In \textit{Test-2}, MedSAM produces incomplete masks for fine defects, while SAM-Med2D introduces spurious detections. In contrast, XCT-SAM generates spatially accurate segmentations that better preserve defect structures. In \textit{Test-5}, XCT-SAM also correctly segments sparse pores without false merging. Overall, XCT-SAM produces predictions that more closely align with reference masks across datasets, defect types, and imaging conditions.


\section{Conclusions}
We present XCT-SAM, a sequential parameter-efficient domain adaptation framework for defect segmentation in AM XCT images. In the first stage, we adapt Conv-LoRA modules on alloy microstructure images using a lightweight configuration with rank r=2 and eight convolutional experts, while keeping over 99\% of SAM parameters frozen. In the second stage, we further adapt the model using GAN-generated XCT data, progressively bridging the gap between natural-image pre-training and XCT defect segmentation. Experimental results show that XCT-SAM outperforms strong baselines such as UNet++, MedSAM, SAM-Med2D, and direct Conv-LoRA-SAM on synthetic OoD and real NIST XCT benchmarks, particularly in IoU and Dice performance. These results suggest that intermediate material-domain adaptation can improve the generalization of parameter-efficient foundation-model adaptation under limited and imbalanced XCT supervision. Current limitations include the additional cost of two-stage training, reduced robustness under severe noise and contrast shifts, and reliance on 2D slice-wise binary models. Future work will explore continual cross-material adaptation, full 3D volumetric modeling, and unified multiclass defect segmentation. Overall, XCT-SAM provides a scalable framework for AM XCT defect segmentation in data-scarce settings.


\section*{Acknowledgment}
\noindent This work was supported by a gift from Qualcomm Incorporated. The authors would also like to acknowledge the Pacific Research Platform, NSF Project ACI-1541349, and Larry Smarr (PI, Calit2 at UCSD) for providing the computing infrastructure used in some of the experiments.

%
%
\bibliographystyle{IEEEtran}
\bibliography{main.bib}
\end{document}